# Standard methods for inexpensive pollen loads authentication by means of computer vision and machine learning

**Manuel Chica[1,2], Pascual Campoy[3]**

**[1]European Centre for Soft Computing, Gonzalo Gutiérrez Quirós, 33600 Mieres (Asturias), Spain**

**[2]Inspiralia Tecnologías Avanzadas, Estrada 10, 28034 Madrid, Spain**

**[3]Automatics and Robotics Center - Universidad Politécnica de Madrid, José Gutiérrez Abascal 2, 28006 Madrid, Spain**

**Corresponding author: manuel.chica@softcomputing.es**

## Table of contents

## Summary

We present a complete methodology for authenticating local bee pollen against fraudulent samples using image processing and machine learning techniques. The proposed standard methods do not need expensive equipment such as advanced microscopes and can be used for a preliminary fast rejection of unknown pollen types. The system is able to rapidly reject the non-local pollen samples with inexpensive hardware and without the need to send the product to the laboratory. Methods are based on the color properties of bee pollen loads images and the use of one-class classifiers which are appropriate to reject unknown pollen samples when there is limited data about them. The validation of the method is carried out by authenticating Spanish bee pollen types. Experimentation shows that the proposed methods can obtain an overall authentication accuracy of 94%. We finally illustrate the user interaction with the software in some practical cases by showing the developed application prototype.

**Short title**

Automatic pollen loads authentication

# 1. Introduction

Bee-keepers, bee-keeping associations, and laboratories are interested in detecting fraud in pollen, and require tools to standardize and authenticate bee pollen origin in order to guarantee their nutritive and health benefits. Microscopic analysis of pollen grains, which form bee pollen loads, is a precise method of identifying origin. However, this process requires the laboratory work of melissopalynology experts, and is thus time consuming and costly.

There have been many attempts to automate pollen grain identification by computer algorithms but there is no inexpensive, complete, and automated process (Allen 2006, Boucher et al. 2002, Rodríguez-Damián et al. 2006). Additionally, experts use macroscopic identification of bee pollen loads by means of such properties as color. This method, although it cannot guarantee complete accuracy, can provide an initial, reliable idea of bee pollen origin (Kirk 1994). Also, some melissopalynology experts separate pollen load samples by color as a previous step to final microscopic identification (Campos et al. 1997, M. P. de Sá Otero 2002). This process is carried out manually by experts who spend more than an hour separating each sample, an indication of the difficulty and subjectivity involved. Thus, the development of a completely automated, inexpensive system which can recognize external pollen load properties, such as color, can bring about a two-fold improvement in bee pollen origin authentication: a) recognition of local bee pollen by non-experts (bee-keeping associations, for instance) and b) reduction of laboratory work by experts through separating the pollen loads automatically.

The development of an automatic system to separate and recognize the pollen loads types by color is a complex task. The main computational problems are two:

a) even within a single pollen type, color variability is high due to environmental characteristics (humidity during the plant growth, drying process, or the presence of impurities); and b) the classification of known local pollen loads must be made against all other world pollen types (initially unknown). This is an important obstacle for designing an automated system since color data cannot be collected from all existing bee pollen types. In order to overcome these difficulties a preliminary computational proposal to identify fraudulent pollen types is given in Chica and Campoy (2012).

The latter methodology used image processing techniques and one-class classification algorithms to identify unknown pollen types. Concretely, the mean shift algorithm (Comaniciu and Meer 2002) filters the pollen image and is used to homogenize pollen load color information. Then, one-class classification algorithms (Moya et al. 1993, Chandola et al. 2009) identify each local pollen type. The method makes use of a multi-classifier algorithm, designed to aggregate one-class classifier outputs, given a unique response with a confidence measure. Finally, an ambiguity discovery algorithm is also included to detect identical pollen type and reduce misclassification.

In this contribution we will review the details of the algorithms involved in the methodology, describe the necessary hardware equipment for the standard method, and present the results of applying the image processing algorithms and multi-classifiers. In addition, we present the overall process to create a software dictionary with known local pollen types, use the graphical user interface, and separate and analyze the pollen loads of the images by type. The validation of the proposed standard methodology is carried out for the authentication of four of the most common Spanish pollen types, *Cistus ladanifer*, *Rubus, Echium*, and *Quercus ilex*, with respect to non-Spanish pollen types (M. P. de Sá Otero 2002). Totally, a dataset of around 2000 instances has been used to validate the trained system.

The remainder of the paper is structured as follows. In Section 2 we give a review of the existing image processing and machine learning contributions. Then, Section 3 details the specifications of the hardware necessary for running the methods. The algorithms of the methodology, i.e. image processing and one-class multi-classifiers, are described in Section 4. We present the image processing results, classifiers performance, and validation of the process by using the software prototype in Section 5. Finally, in Section 6, we highlight some concluding remarks and future works.

## 2. Review on computer vision and classification methods for pollen recognition

Although there are limited discernment methods for recognizing pollen types in macroscopic images (Carrión et al. 2004, Chica and Campoy 2012) the majority of the existing methods for analyzing bee pollen and its origin are applied to microscopic pollen grains images. Some systems use scanning electron microscopy (SEM) images (Treloar et al. 2004). There are also systems based on laser scanning (Ronneberger et al. 2002).

The first works on recognizing pollen grains by optical microscopes were presented by France et al. (2000) and Boucher et al. (2002) where some discriminative features of various pollen taxa were detected and classified. Then, Li et al. (2004) and Zhang et al. (2004) extracted more sophisticated information from pollen grains such as Gabor wavelets and moment invariants. They also implemented an artificial neural network for classifying pollen grains. Rodríguez-Damián et al. (2006) obtained an accuracy of 89% while classifying similar species of the *Urticaceae* family using shape and texture features, neural networks, and support vector machines. An additional feasibility study on recognizing the pore and colpi structures of grass, birch, and mugwort pollen grains is done by Chen et al. (2006). Landsmeer et al. (2009) propose a mechanism to identify pollen grains on air samples against other microscopic particles.

Pollen authentication is a more specific problem in literature where there is limited data to model the non-local pollen types (negative classes or outliers). Although it is possible to model the local pollen types it is not possible to model all the existing fraudulent pollen types from around the world. One-class classification is an appropriate machine learning paradigm to deal with this problem (Moya et al. 1993, Chandola et al. 2009). Some authors have used these techniques to identify fraudulent samples in macroscopic (Chica and Campoy 2012) and microscopic pollen images (Chica 2012).

One-class classification is different from conventional binary or multi-class classification. This distinction lies in the absence of the negative class (normally called outlier) or in the vagueness of its definition and sampling (Chandola et al. 2009). Originally, the term was given by Moya et al. (1993) and some authors refer to this problem as outlier detection (Ritter and Gallegos 1997) or novelty detection (Bishop 1994).

# 3. General overview of the proposed method and hardware equipment

Diagram of Fig. 1 shows the outline of the proposed method. Initially, the experimenter prepares the pollen loads sample to acquire the image by the computer vision system. The pollen loads of the image are segmented from background (see Section 4.1) and filtered (see Section 4.2). Then, the color instances of the processed pollen loads are used to train a multi-classifier model based on one-class classifiers (one for each local pollen type). Finally, the multi-classifier outputs the authentication of each color instance, classifying them as a known local or non-local (outlier) pollen type (see Section 4.3).

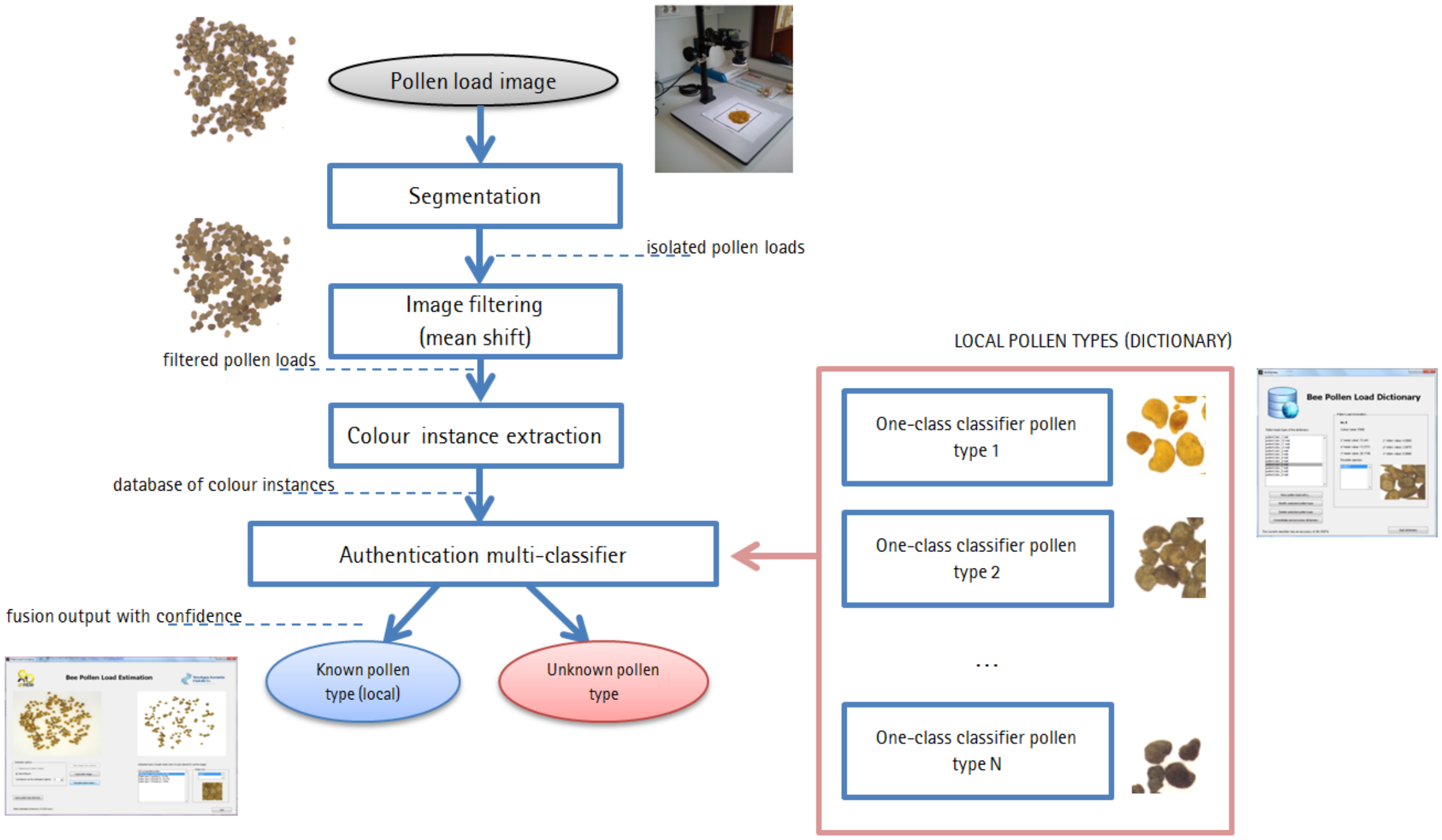

***Fig. 1:*** The diagram shows the proposed standard method for detecting unknown pollen loads. The processing chain finishes with an authentication output.

Regarding the equipment for using the methodology it is important to remark that the time saving of the presented approach is high since, without an automatized method, experts do usually need hours to manually separate the pollen loads. Moreover, the cost of the needed equipment is low. Mainly, the set of components for the vision system comprises:

- **Camera and optics device**: to obtain images of the pollen load samples. The resolution does not need to be extremely high since only color and edge information will be processed. In our case, we have selected the color camera uEye UI-1485LE-C (www.ids-imaging.com) with an Aptina CMOS sensor in 5 MPixels resolution (2560x1920 pixels). The light-weight housing of the UI-1485LE features a C/CS lens mount with adjustable flange back distance. The approximate cost of the camera is 500€. Additionally, and mounted on the electronic camera, we need a focal lens (obtained from Goyo Optical (www.goyooptical.com) with a cost of 150€.

- **Lighting conditions**: in almost all the computer vision applications, the lighting conditions are decisive and must be controlled. We have considered an external lighting generator which points out the pollen load sample. In this way, the color and pollen loads characteristics will not vary during the different runs. The external lighting consists of a 50 LEDs based ring light from CCS Direct Lighting (www.ccsgrp.com). It can be mounted on a CS mount and its cost is around 450€.

- **Support device:** we need a supporting device to allow a fixed position for the camera, lens, and lighting. The cost of this support is almost inexistent, with models from 50€. The sample of pollen loads is placed on the base of this supporting device, getting a stable and low-cost system to acquire images of the sample to a posterior processing and analysis (see Fig. 2).

- **Computer:** a domestic computer (around 800€) must be connected to the camera via USB to install the software. Proprietary software can be used to take snapshots of the pollen load samples. Then, the vision system will be controlled by the software prototype to ease the labor of the experimenter.

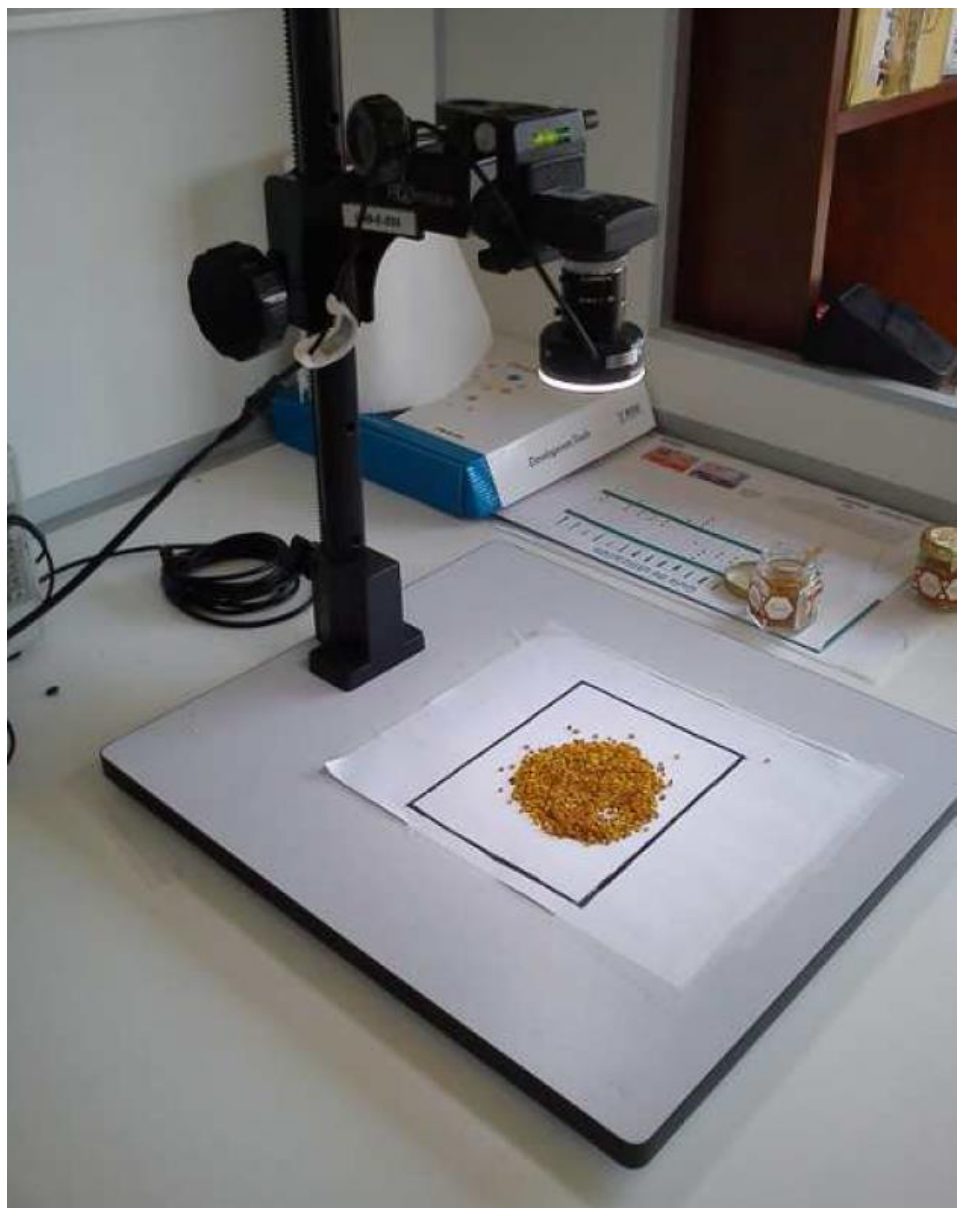

***Fig. 2:*** The computer vision system needs of inexpensive hardware such as a camera and a support to control illumination.

## 4. Description of the computational techniques needed for the methodology

### 4.1 Segmentation algorithm

First, the well-known Otsu segmentation algorithm (Otsu 1979) is applied to the gray-scale image to extract the pollen loads from the background. Then, a morphological opening operation is applied to the thresholded binary image (Gonzalez and Woods 2008). This operation removes those small objects having less than 50 connected pixels in an 8-connected neighborhood.

Pixels extracted in the latter phase are to be analyzed by the remaining processing algorithms. Their color information can be represented in several ways. The most common is the RGB space where colors are represented by their red, green, and blue components in an orthogonal Cartesian space. However, the RGB space does not lend itself to mimicking the higher level processes which allow human color perception.

Color is better represented in terms of hue, saturation, and intensity, as HSI or HSV spaces do (Lucchese and Mitra 2001). However, the latter color spaces are not perceptually uniform. The CIE and $L^*u^*v^*$ and $L^*a^*b^*$ are ideal for color recognition because of the following three properties: a) separation of achromatic information from chromatic information, b) uniform color space, and c) similarity to human visual perception. In these color spaces, for instance, the Euclidean distance between two color points can be easily calculated. This property will ease the work of the classification algorithms.

### 4.2. Homogenizing the pollen loads of the image by mean shift filtering

Each extracted pollen load from the image has many different $L^{*}u^{*}v^{*}$ color values, possibly as many as pixels contained in the load. This has a negative impact on pollen load color authentication, since human experts customarily identify each pollen load as a unique color. Then, it is necessary to apply an image processing algorithm before using the classification methods. The goal is to homogenize and smooth the large quantity of different color points of a pollen load in just a few unique and representative color instances.

One of the best methods for discontinuity-preserving smoothening in image processing is the mean shift algorithm, proposed in Comaniciu and Meer (2002), and in line with the feature space analysis. The main strengths of the mean shift algorithm are: a) it is an application independent tool, b) it is suitable for real data analysis, c) it does not assume any prior shape on data clusters, d) it can handle arbitrary feature spaces, and e) it has only one parameter, the bandwidth selection window size, to be chosen.

The mean shift procedure, originally presented by Fukunaga and Hostetler (1975), is a procedure for locating the maxima of a density function given discrete data sampled from that function. It is also an iterative method, starting with an initial estimate or points $x_i$ and a $G$ kernel function. This kernel function determines the weight of nearby points for the re-estimation of the mean. Denote by $y_j$ with $j = 1,2,\ldots$ the sequence of successive locations of the kernel $G$:

$$y_{j+1} = \frac{\sum_{i=1}^{n} x_i G(\|\frac{x-x_i}{h}\|)^2}{\sum_{i=1}^{n} G(\|\frac{x-x_i}{h}\|)^2} \qquad \text{(Equation 1)}$$

is the weighted mean at $y_j$ computed with the kernel $G$ and $y_1$ is the centre of the initial position of the kernel.

Taking this theory as a base, Comaniciu and Meer (2002) extended the mean shift procedure for filtering and segmenting images. A color image is typically represented as a two-dimensional lattice of 3-dimensional vectors (pixels). The space of the lattice is known as the spatial domain, while the color is the range domain. For both, the mean shift algorithm will work with the Euclidean metric defined in the previous section for the color space. When both domains are concatenated, the dimensions of the joint spatial range domain have to be compensated by proper normalization. Thus, the multivariate kernel is defined as the product of two radially symmetric kernels, and the Euclidean metric allows a single bandwidth parameter for each domain:

$$K_{h_s,h_r}(x) = \frac{C}{h_s^2 h_r^p} k(\| \frac{x^s}{h_s} \|^2) k(\| \frac{x^r}{h_r} \|^2), \qquad \text{(Equation 2)}$$

where $x^s$ is the spatial part, $x^r$ is the range part of a feature vector, $k(x)$ the common profile used in both domains, $h_s$ and $h_r$ the kernel bandwidths, and $C$ the normalization constant. As the normal kernel always provides satisfactory results, the user just has to provide one parameter $h = (h_s, h_r)$ which controls the size of the kernel, and then the smoothening resolution. Nevertheless, replacing the pixel in the centre of the window by the average of the pixels in the window blurs the image. Discontinuity-preserving smoothening techniques, on the other hand, reduce the amount of smoothening near abrupt changes. The mean shift algorithm uses a bilateral filtering which works in the joint spatial-range domain. The data are independently weighted in the two domains and the centre pixel is computed as the weighted average of the window. The kernel in the mean shift procedure moves toward the maximum increase in joint density gradient, while bilateral filtering uses a fixed static window.

To sum up, the mean shift filtering algorithm works as follows. Let $x_i$ and $z_i$, with $i = 1,2,\dots,n$ be the (p+2)-dimensional input and filtered image pixels in the joint spatial-range domain. Being the superscripts $s$ and $r$ the spatial and range components of a vector, respectively, and $c$ the point of convergence, for each pixel:

1. Initialize $j = 1$ and $y_{i,1} = \mathrm{x_i}$.

2. Compute $y_{i,j+1}$ according to Equation 1 until convergence, $y = y_{i,c}$

3. Assign $z_i = (x_i^s, y_{i,c}^r)$.

The spatial bandwidth has a distinct effect on the output when compared to the range (color) bandwidth. Only features with large spatial support are represented in the filtered image when $h_s$ increases. On the other hand, only features with high color contrast survive when $h_s$ is large.

#### 4.3. One-class multi-classification algorithm based on distances

In Chica and Campoy (2012) authors compared four different one-class classification approaches for the problem: Gaussian, Parzen classifier, SVDD and $kNN$. The latter algorithm, $kNN$ with $k = 1$, showed the best performance. Then, we will use this algorithm for the pollen recognition system.

$kNN$, originally provided by Dasarathy (1991), is a distance-based one-class classifier based on the assumption that normal data instances occur in dense neighborhoods while anomalies occur far from their closest neighbors. The basics of the algorithm for one-class classification are that the anomaly score of a data instance is defined as the distance with its $k^{th}$ nearest neighbor in a given dataset.

Nearest neighbor classifiers always require the definition of distance or similarity measures between two data instances. For continuous features, the Euclidean distance is the most popular choice. In the case of choosing $k = 1$ as the parameter of the algorithm each new instance $z$ will be considered as target or outlier depending on the classification of its closest neighbor in the training data.

In multi-class anomaly detection (our pollen authentication problem) training data contains labeled instances belonging to multiple normal classes but it does not contain anomalous instances. A test instance is considered anomalous if it is not classified as normal by any of the classifiers. To do this, a confidence score with the prediction made by the classifier is normally provided. If none of the $kNN$ classifiers are confident in classifying the test instance, the instance is then declared to be anomalous (Chandola et al. 2009).

We have followed the latter approach, modeled as follows: $|C|$ being a set of known local bee pollen load types, the training data will contain instances belonging to $|C|$ classes. In order to use one-class $kNN$ classifiers and be able to reject unknown pollen load types, we have to decompose the classification system in $|C|$ binary sub-problems. Thus, we will train $|C|$ different $kNN$ classifiers: $f_1, f_2, \dots f_{|C|}$. An ensemble scheme is used to fuse all of them in a multi-class authentication output.

Therefore, for each pollen color instance $x$ we first map each one-class classifier output $f_i(x)$ to a posterior probability $P(y = c|x)$. These probabilities are also normalized in the range $[0,1]$. The posterior probability of each classifier's target can be considered as the confidence $CF_{oc}(y = c|x)$ for one instance $x$ to belong to class $c$.

In order to classify an incoming pollen load sample as one of the $|C|$ possible pollen types we build a multi-classifier. The multi-classifier will compare the confidence $CF_{oc}(y|x)$ of all the one-class classifiers and will provide with a global prediction from the most reliable one-class classifier. The multi-classifier prediction $\omega$ is given by:

$$\omega = max_{1 \le c \le |C|} CF_{oc}(c|x). \qquad \text{(Equation 3)}$$

But it is also necessary to estimate the confidence of the multi-classifier prediction. To do this we first introduce two parameters, $T_{oc}$ and $T_m$, as done in Goh et al. (2005):

$$T_{oc} = CF_{oc}(\omega|x), \qquad \text{(Equation 4)}$$

$$T_m = T_{oc} - max_{1 \le c \le |C|, c \ne \omega} CF_{oc}(c|x). \qquad \text{(Equation 5)}$$

$T_{oc}$ is the highest confidence factor from the $|C|$ binary one-class classifiers and determines the multi-classifier prediction class $\omega$, $T_{oc}$ might not be sufficient to estimate the global confidence of the multi-classifier prediction. This is the reason for introducing the second parameter: the multi-class margin $T_m$. Wrong predictions could have high $T_{oc}$ but small $T_m$ ; correct predictions must have higher multi-class margin values $T_m$.

Goh et al. 2001 showed that there is a better separation of correct from erroneous predictions if the multi-class margin variable is used. After a preliminary experimentation, we set parameters $T_{oc}$ and $T_m$ to 0.5 and 0.01, respectively, to be used in the final decision of the multi-classifier as in the rule of Equation 6.

$$\begin{cases} \omega \text{ is accepted,} & \text{if } T_{oc} \ge 0.5 \text{ and } T_m \ge 0.001, \\ \text{outlier,} & \text{otherwise.} \end{cases} \qquad \text{(Equation 6)}$$

Sometimes one or more bee pollen types could have exactly the same color description as another. In that case, the $kNN$ multi-classification system must be able to detect, during the training phase, that one incoming local pollen type is identical to one already existing (i.e. it is already included in the dictionary) and, at least, warm the user of the software. This mechanism is called ambiguity discovery. Mathematically, $\varepsilon_{m_1}$being a sensitivity-specificity error of a multi-classifier before the inclusion of the new pollen type, and $\varepsilon_{m_2}$the error of a multi-classifier after the inclusion of the classifier of the new pollen type, $\Delta_\varepsilon = (\varepsilon_{m_2} - \varepsilon_{m_1})$ is the difference between them. The ambiguity discovery process is launched every time the $\Delta_\varepsilon$ parameter is higher than a fixed value.

In our case we have used the F-measure (van Rijsbergen 1979) as the $\varepsilon_{m_1}$error measure. Where $\Delta_\varepsilon$ exceeds a threshold value, ambiguity discovery is triggered and the process works as follows:

1. The confusion matrix of the new multi-classifier for the testing data is computed.

2. The maximum value of $(\Lambda_i, L_j)$ is calculated with $i \neq j$, $\Lambda$ being the vector of real classes and $L$ the vector of predicted classes.

3. The user is consulted about merging conflicting classes $c_i$ and $c_j$ into a unique class $c_i$.

4. The multi-classifier is trained according to the response of the user in the third step.

## 5. Validation results and discussion

We first describe the data used for experimentation and the image filtering results (Section 5.1). By using the processed color data from pollen images we show and analyze the one-class classifiers and multi-classifier results (Section 5.2). Finally we show the software prototype used to validate the system as an illustrative example of the proposed standard process (Section 5.3).

### 5.1. Experimental data and image processing results

Different samples of Spanish bee pollen loads were obtained from beekeepers to build the authentication models and validate them against nonlocal samples. Samples belonging to the four local pollen types (*Rubus, Echium, Cistus ladanifer*, and *Quercus ilex*) and non-local samples were identified and grouped by experts. Images where these pollen loads samples appear are in Fig. 3. In this figure it can be observed how, even for experts, color separation is difficult and subjective, and how non-local samples can be misleading (image on the right of the figure).

Processing these images of the pollen samples (taken in TIFF format and resolution of 1024 x 768 pixels) is the first step of the method. The software will apply the filtering algorithm after segmenting pollen loads from background. As explained in Section 4.2, the selection of an appropriate bandwidth parameter for the mean shift algorithm is not trivial. This selection depends on the kind of images used in the application. Basically, the implemented mean shift algorithm receives the spatial bandwidth $h_s$, the range bandwidth $h_s$, and the minimum segment area in number of pixels. The last parameter is fixed to a high value (20 pixels) in order not to segment a pollen load into different parts.

The selection of the other two parameters is more difficult. High spatial bandwidths merge different bee pollen loads because they are normally close to each other. Low range bandwidths do not effectively aggregate the entire color information of the pollen loads. We conducted a preliminary experiment with different values and, to illustrate the importance of this parameter selection, we show in Fig. 4 an original *Rubus* data sample with three output images after applying the filtering algorithm with $h_s$ to 7, 15, and 30, respectively. Additionally, it is important to remark that higher bandwidth values mean higher computational time. We set the bandwidths of the algorithm to $h = (h_s, h_r) = (15{,}20)$ for the remaining experimentation which seems to be one of the best parameter combination for the pollen load problem.

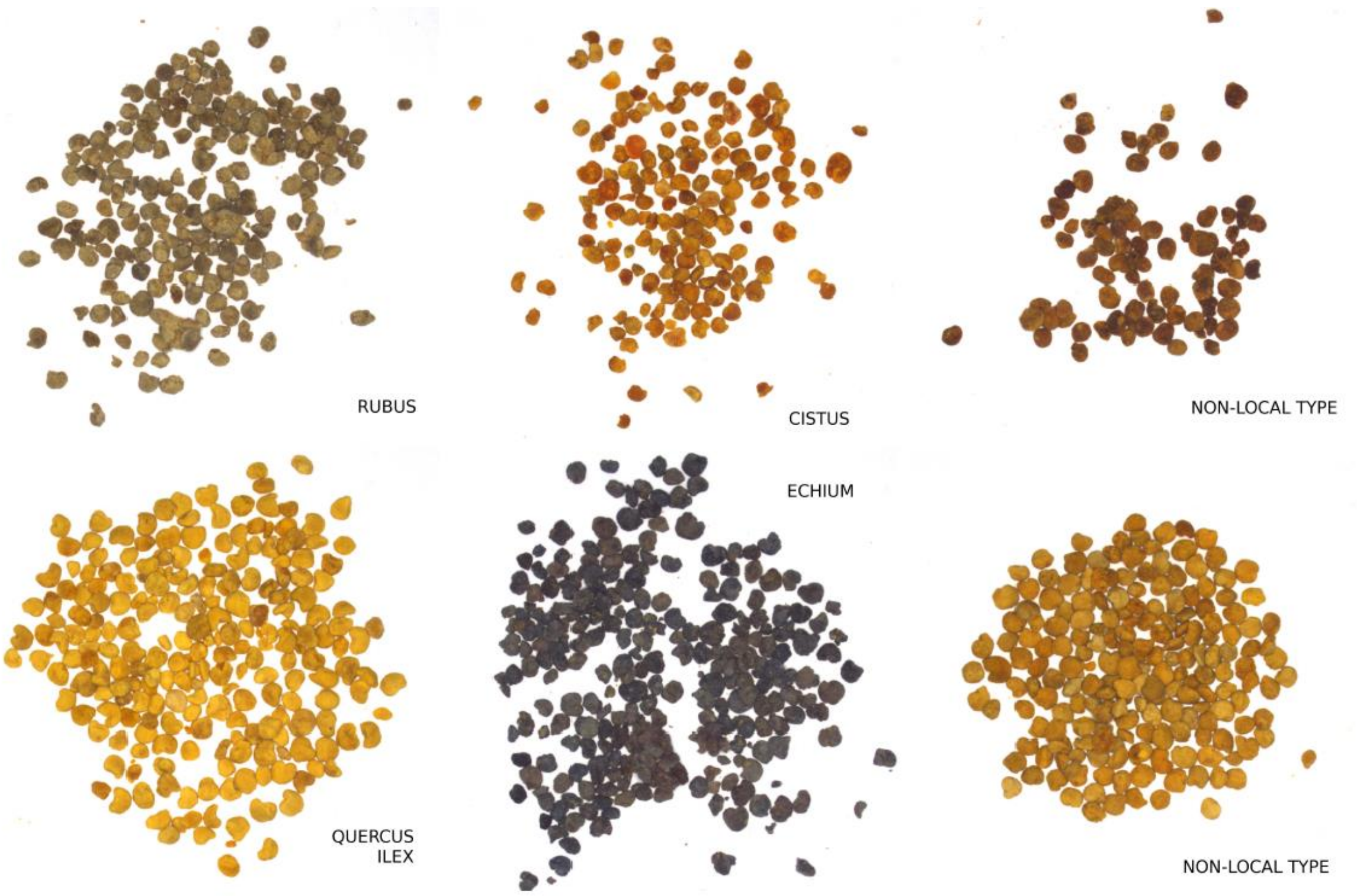

***Fig. 3:*** Different images of pollen load samples which were taken with our vision system. Left and central images belong to known pollen types (*Rubus, Cistus ladanifer, Quercus ilex*, and *Echium*, respectively). Right images are non-local samples and must be rejected by the system.

From the processed and filtered images we generated a set of 3146 color instances, each with three input features (corresponding to the $L^{*}u^{*}v^{*}$ color space values) and its class (one of the four pollen types or an outlier class). Of this set of color instances, 400 are used for the training of the four one-class classifiers (100 each pollen type), 800 to test the one-class classifiers (400 belonging to the four known pollen types and 400 were outliers), and the remaining 1946 instances to validate the multi-classification system (Section 5.2). The latter three sets of instances are independent. We have summarized the experimental data in Table 1.

**5.2. Numerical performance of the multi-classification algorithms**

We have calculated the classification accuracy, false negative and positive rates, F-measure, and confusion matrix in order to numerically validate the multi-classifier performance. A false negative (FN) occurs when the outcome of the classifier is incorrectly predicted as outlier when it is actually a target. A false positive (FP), on the other hand, occurs when the outcome is incorrectly predicted as target when it is actually an outlier. The FN rate measures the number of FNs out of the total number of negatives or outliers, and the FP rate calculates the fraction of FPs divided by the total number of positives or target instances (Witten and Frank 2005). The F-measure provides a relation between the precision and recall of the classification results (van Rijsbergen 1979).

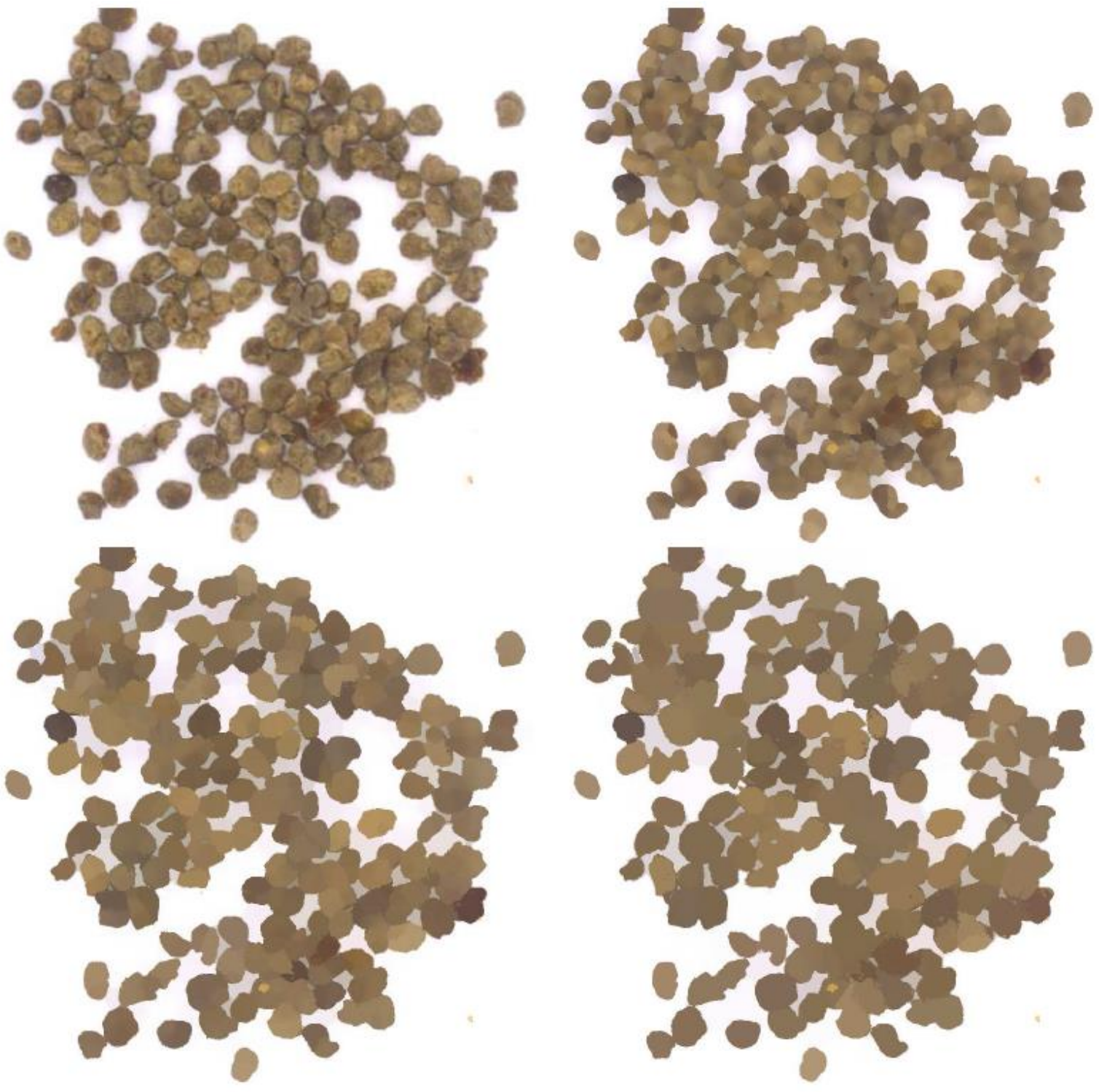

***Fig. 4:*** Original image and resulting images after applying the mean shift algorithm (varying spatial bandwidth $h_s$ to 7, 15, and 30, respectively).

First, we have analyzed the performance of the $kNN$ classification method for the authentication of each of the selected local bee pollen types in isolation. These results were obtained by classifying the test dataset with the trained one-class classifiers. The classifiers were trained without rejecting any training instance as outlier (rejection threshold equal to 0%). Table 2 shows the evaluation measures of the classifiers for each pollen type.

**Table 1:** Description of the three independent datasets used to train, test the classifiers, and validate the one-class multi-classifier.

| Dataset name | Total size | Pollen types instances | Outlier instances |
|---|---|---|---|
| Training | 400 | 400 | 0 |
| Test | 800 | 400 | 400 |
| Validation | 1946 | 1016 | 930 |

**Table 2:** Evaluation measures obtained by the one-class $kNN$ classifiers for each of the four pollen types. Lower values mean better performance.

| ***Rubus*** | | | ***Echium*** | | |
|---|---|---|---|---|---|
| FN rate | FP rate | F-measure | FN rate | FP rate | F-measure |
| 0.05 | 0 | 0.9744 | 0.09 | 0 | 0.9529 |
| ***Cistus ladanifer*** | | | ***Quercus Ilex*** | | |
| FN rate | FP rate | F-measure | FN rate | FP-rate | F-measure |
| 0.06 | 0 | 0.9691 | 0.01 | 0.0028 | 0.9851 |

By observing figures of Table 2 we can conclude that the FP rate is almost 0. This means that the classifiers are able to correctly identify all the outliers (non-local pollen types) without misclassifying them as local pollen types. The FN rate is higher than the FP rate although its value is low, 9% in the worst case (*Echium* pollen type).

After showing the results of the isolated $kNN$ classifiers we also present the results of the multi-classifier method formed by the four latter $kNN$ classifiers. We run the multi-classifier under two different training conditions. The first condition is defined to use one-class classifiers having a 0% threshold rejection. The second one is devoted to use a 10% rejection. Results are calculated by applying the multi-classifiers to the validation dataset, totally independent of the training and test datasets.

Table 3 shows the performance measures of the multi-classifier. In the first block of figures we have shown the accuracy and FP-FN rates when rejection threshold equals 0%. In the second block, a rejection threshold of 10% is used for obtaining the results. The $kNN$, with a rejection of 10%, is the model with the best results as it has low FP and FN rates, and also the highest accuracy.

**Table 3:** Evaluation measures of the final multi-classifier. Two cases are listed; considering 0%, and considering 10% of data as outliers when training the one-class classifiers. FN and FP rates are calculated taking all the known pollen types as the positive class and the outliers as the negative.

| $kNN$ **multi-classifier model** | **Accuracy (%)** | **FN rate** | **FP rate** |
|---|---|---|---|
| *0% rejected during training* | 92.5488 | 0.0108 | 0.1161 |
| *10% rejected during training* | 94.6043 | 0.0935 | 0.0167 |

Table 4 shows the confusion matrix of the best multi-classifier, $kNN$ with a rejection of 10%. It can be seen that generally there is no miss-classification between the known pollen load types, just one instance in which the $kNN$ multi-classifier is misclassified. The highest number of errors is located in the right column. These are FNs as real pollen load type instances are classified as outliers (non-local pollen types). Nevertheless, as can be seen in Table 3, the FN rate is not high.

We can also see that the trade-off between FP and FN rates is better than in the isolated one-class classifiers. This fact could be justified because of the confidence mechanism of the multi-classifier which is able to discard instances classified by one of the one-class classifiers as local pollen types when they are not clearly confident about their decision. Although accuracy is obtained when classifying all the classes (four known pollen types and outliers), the FP and FN rates are calculated between the classification of the instance as known pollen type (one of the four known types) or as outlier. There is almost no error among local pollen types.

Finally, if we compare the validation measures of the multi-classifiers and one-class classifiers we can observe how the overall results of the multi-classifiers are better than the isolated one-class classifiers. This fact indicates good behavior of the fusion scheme of the known pollen type classifiers.

**Table 4:** Confusion matrix of the multi-classification system formed by the four one-class kNN classifiers (one for each pollen type). 10% of the training data were rejected as outliers during the training phase of the one-class classifier.

| | | Predicted pollen type | | | | | |
|---|---|---|---|---|---|---|---|
| | | ***Rubus*** | ***Echium*** | ***Cistus*** | ***Quercus*** | **Outlier** | **Total** |
| **Real pollen type** | ***Rubus*** | **248** | 0 | 0 | 0 | 33 | 281 |
| | ***Echium*** | 0 | **357** | 0 | 0 | 43 | 400 |
| | ***Cistus*** | 0 | 0 | **275** | 1 | 10 | 286 |
| | ***Quercus*** | 0 | 0 | 0 | **48** | 1 | 49 |
| | **Outlier** | 4 | 0 | 5 | 8 | **913** | 930 |
| | **Total** | 252 | 357 | 280 | 57 | 1000 | **1946** |

#### 5.3. Validation of the complete prototype

We developed a software prototype containing the validated methods to manage the created local pollen types, train the models, and validate the whole process. The software prototype was programmed in MATLAB using some functions of the DD tools library (Tax 2011). Initially, the experimenter needs to train the system with the known local samples. She/he must manually separate pollen types as shown in Fig. 5. By using the dictionary tool of the software prototype the experimenter can save her/his local (known) types. See for instance Fig. 6 where an entry called *Rubus* is introduced into the system.

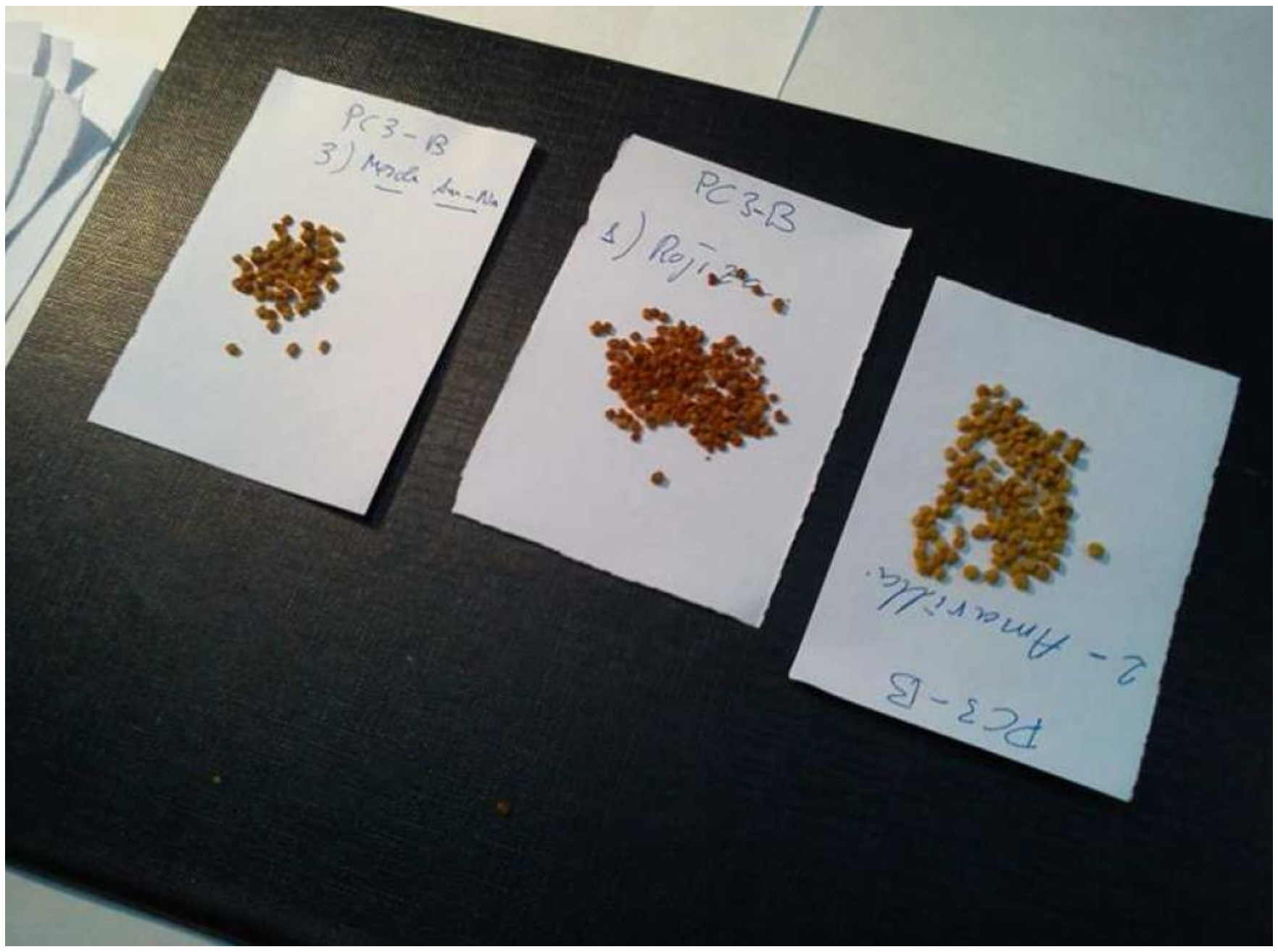

***Fig. 5:*** Pollen load samples separated by an expert to train the system.

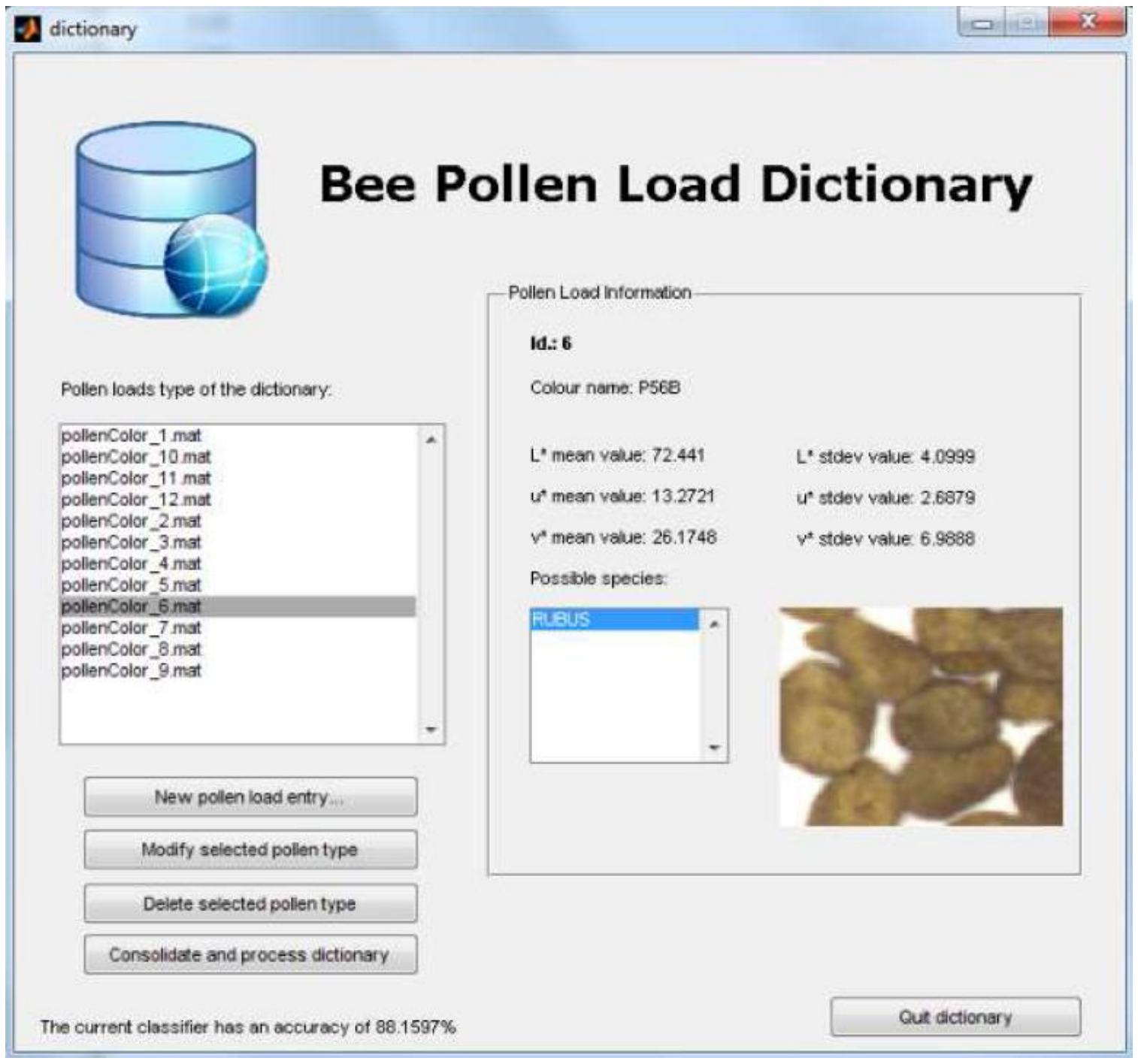

***Fig. 6:*** Screenshot of the pollen dictionary of the prototype. The ambiguity discovery algorithm checks if there is an existing pollen type having identical features with respect to an incoming one.

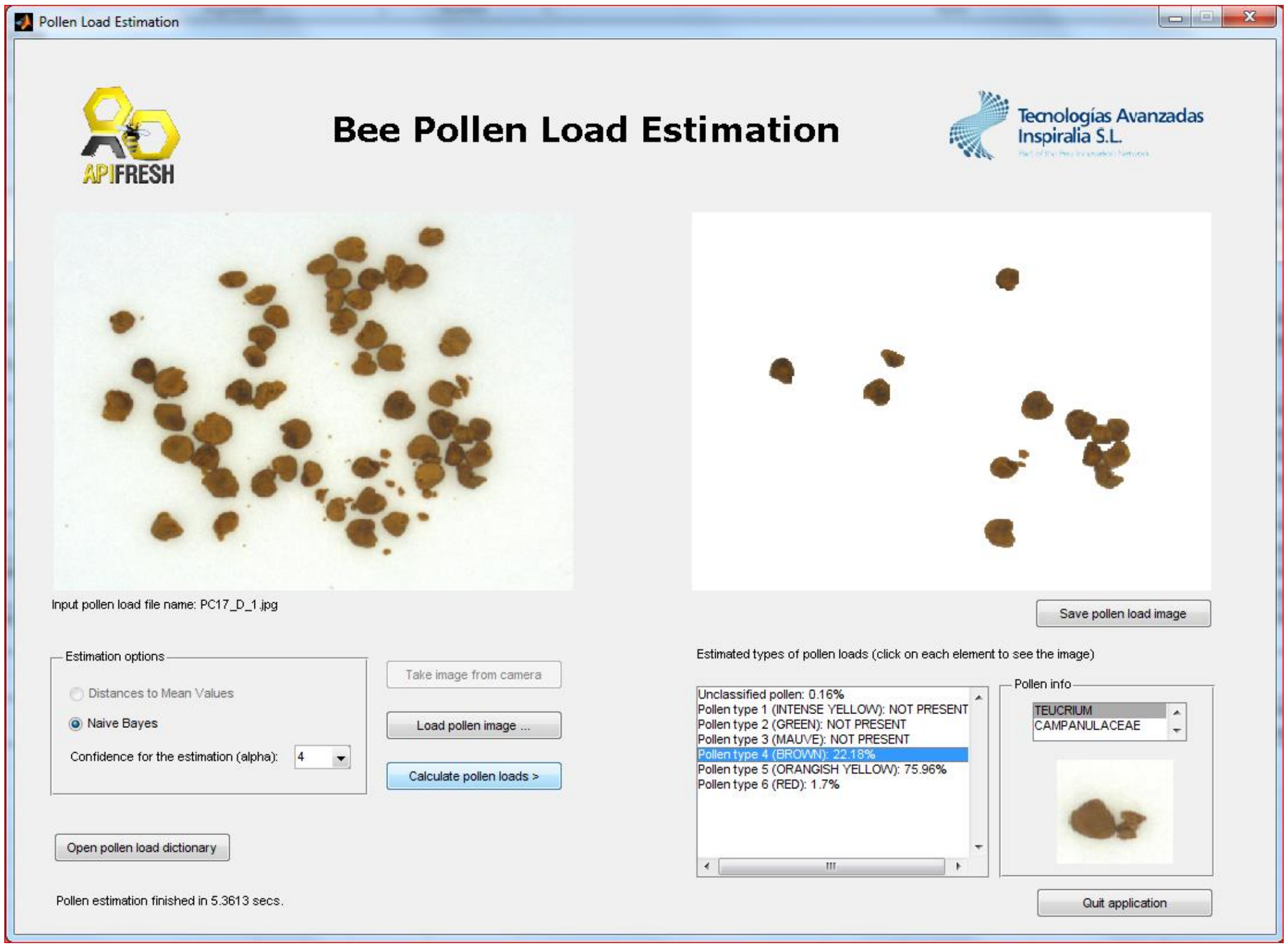

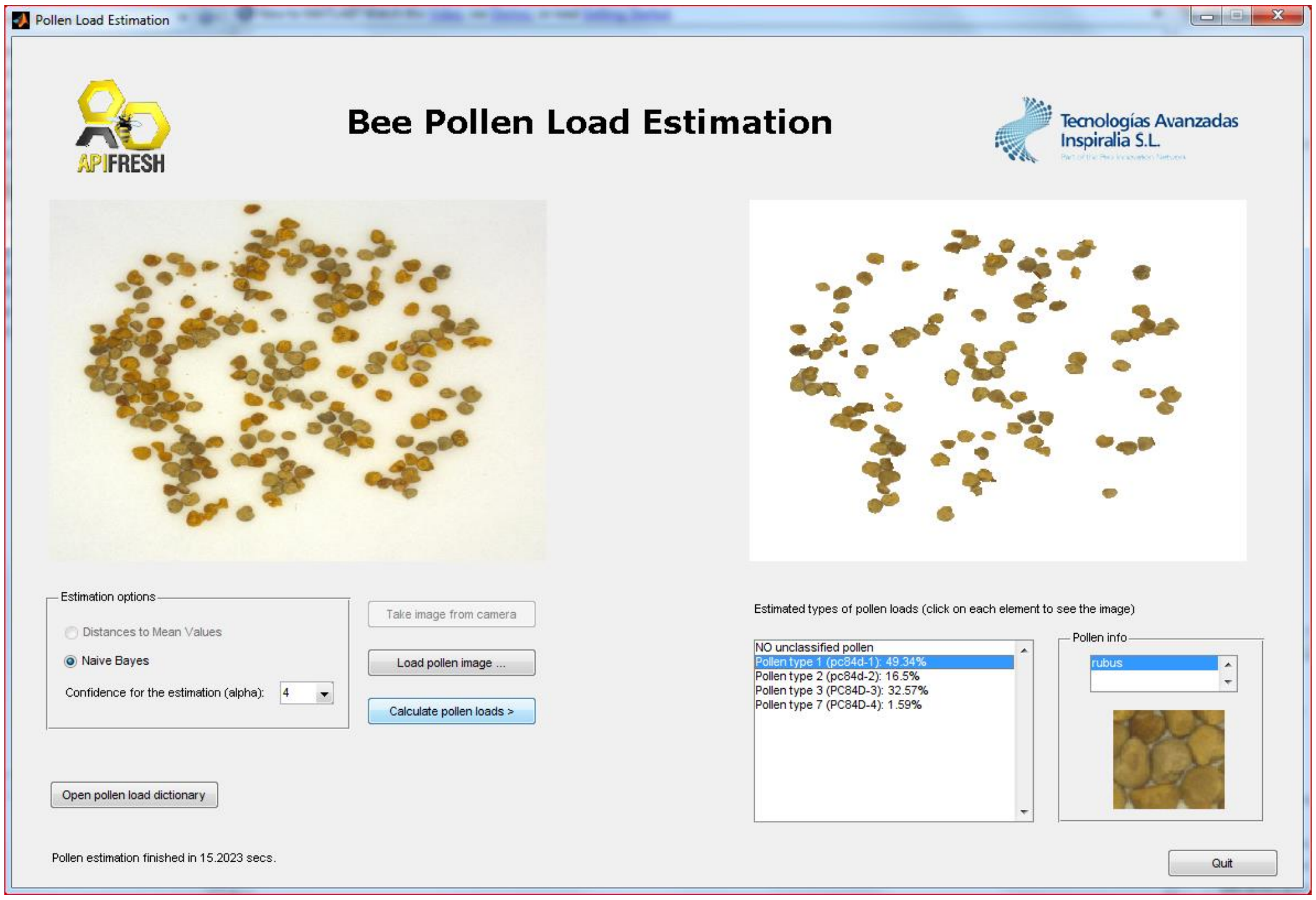

**Fig. 7:** Two screenshots of the software prototype detecting local pollen type in the pollen loads images.

After the training phase of the system, new incoming pollen samples can be introduced in the system to start the automatic pollen loads separation by color. We provide with some application examples in Fig. 7. In these images the graphical user interface shows the list of separated pollen loads by type on the right hand of the screen.

## 6. Conclusions

A review of the existing methods for using computer vision and machine learning for pollen classification is presented at the beginning of the paper. Then, we proposed a standard methodology using a chain of methods based on computer vision and classification techniques for the origin authentication of bee pollen loads against fraudulent samples. The necessary equipment for using the methodology is listed: camera, lens, lighting, mounting and support device, and computer.

We showed the results of applying the image processing algorithms to obtain the color properties of the pollen images and analyzed the impact of selecting appropriate parameters for the mean-shift algorithm. Then, we tested the different one-class classification models based on $kNN$ and the multi-classifier scheme which improves the accuracy of the isolated one-class classification methods and merging the outputs of the classifiers. Therefore, the best multi-classifier formed by $kNN$ one-class classifiers and rejecting some instances in the training phase, has achieved 94% accuracy when detecting local pollen types.

These models have been validated in 1946 colors instances for the authentication of four Spanish pollen types against different outlier samples. We have also shown the complete process of using the application prototype when introducing pollen samples and the results provided by the system.

The use of the presented standard methodology drastically reduce the time and effort spent by experts to several seconds and can be used as an standard method for macroscopically rejecting unknown pollen loads. Future work can be devoted to apply a more interpretable multi-classification system such as fuzzy rule-based classifiers (Alonso and Magdalena 2011). In that way, the users of the standard method could understand the reasons why a sample is rejected as known local pollen type.

## Acknowledgement

The work presented in this paper has been carried out within the scope of the APIFRESH project. APIFRESH has been co-funded by the European Commission under the R4SMEs 7th Framework Program.